\documentclass{article} 
\usepackage{iclr2019_conference,times}

\usepackage{amsmath,amsfonts,bm}

\def\eqref#1{equation~\ref{#1}}

\def\1{\bm{1}}

\DeclareMathAlphabet{\mathsfit}{\encodingdefault}{\sfdefault}{m}{sl}
\SetMathAlphabet{\mathsfit}{bold}{\encodingdefault}{\sfdefault}{bx}{n}

\usepackage{hyperref}
\usepackage{latexsym}
\usepackage{booktabs}       
\usepackage{graphicx}
\usepackage{url}
\usepackage{dirtytalk}
\usepackage{wrapfig}
\usepackage{amssymb}
\usepackage{amsmath}
\usepackage{algorithm}
\usepackage{algpseudocode}

\usepackage{wrapfig}
\usepackage{colortbl} 
\usepackage{caption}
\usepackage{subfig}
\usepackage{array,multirow}

\usepackage{booktabs}       
\usepackage{nicefrac}       

\usepackage{xcolor}
\usepackage{mathtools}

\title{Analysing Dropout and Compounding Errors in Neural Language Models}

\author{
James O' Neill \& Danushka Bollegala  \\
Department of Computer Science\\
Liverpool University\\
Liverpool, L69 3BX, England \\
\texttt{\{james.o-neill, dansuhka.bollegala\}@liverpool.ac.uk} 
}

\begin{document}

\maketitle

\begin{abstract}
This paper carries out an empirical analysis of various dropout techniques for language modelling, such as Bernoulli dropout, Gaussian dropout, Curriculum Dropout, Variational Dropout and Concrete Dropout. Moreover, we propose an extension of variational dropout to concrete dropout and curriculum dropout with varying schedules. We find these extensions to perform well when compared to standard dropout approaches, particularly variational curriculum dropout with a linear schedule. Largest performance increases are made when applying dropout on the decoder layer. Lastly, we analyze where most of the errors occur at test time as a post-analysis step to determine if the well-known problem of compounding errors is apparent and to what end do the proposed methods mitigate this issue for each dataset. We report results on a 2-hidden layer LSTM, GRU and Highway network with embedding dropout, dropout on the gated hidden layers and the output projection layer for each model. We report our results on Penn-TreeBank and WikiText-2 word-level language modelling datasets, where the former reduces the long-tail distribution through preprocessing and one which preserves rare words in the training and test set.

\end{abstract}

\section{Introduction}
Language modelling is a foundational natural language task. In recent times, there has been a surge of research interest in neural language modelling (NLM) which has led to many improvements. This has included a smorgasbord of novel architecture designs ~\cite{merity2016pointer, bradbury2016quasi}, regularization and optimization improvements ~\cite{merity2017regularizing, merity2018analysis} including an evaluation of (1) activation and weight dropping in embeddings, hidden gates in RNNs and output projection layers, (2) the effects of weight tying between input and output layers to reduce parameters in NLMs while improving performance~\cite{inan2016tying}, (3) the effect of varying hidden dimensions of RNNs and embeddings dimensions, (4) optimization techniques such as scheduled cosine annealing of the learning rate with warm restarts ~\cite{loshchilov2016sgdr}, cyclical learning rates where the learning is increased and decreased periodically throughout training~\cite{smith2017cyclical} and separate learning rates for each layer~\cite{singh2015layer} and (5) methods for large scale hyperparameter search for optimal performance~\cite{melis2017state}. Tangentially, there has been methods for addressing salient problems in structured prediction, such as exposure bias~\cite{chang2015learning,norouzi2016reward,bahdanau2016actor,neill2018curriculum} which leads to the compounding of errors is an important concern in language models.

Dropout~\cite{srivastava2014dropout} has been an important regularization technique to prevent overfitting in neural networks by promoting weight sparsity and avoiding weight co-adaptation. Bernoulli and Gaussian Dropout are commonly used in practice. Additionally, for recurrent neural networks (RNNs), fixing the dropout mask over time steps has shown improved performance while being theoretically motivated~\cite{gal2016theoretically}. However, more recent methods in machine learning have looked to learn the probabilities using variational inference methods~\cite{kingma2015variational,gal2017concrete}. However, these novel techniques have been primarily used in the context of other challenges in computer vision and reinforcement learning, but not for natural language tasks. Other dropout techniques such as curriculum dropout~\cite{morerio2017curriculum} require no tuning, instead a schedule that increases dropout throughout training. This can be beneficial when there is some information known about the underlying distribution or the loss space.

Hence, this paper carries out an extensive overview of various dropout techniques and how adapting probabilities can be used to somewhat deter exposure bias. Additionally, we aim to identify commonalities between the networks that benefit the most from each variant and recommend regularization settings when training neural language models. Experiments are carried out on WikiText-2 and Penn Treebank language modelling datasets. Lastly, we analyse the accumulation of errors for a generated sequence of fixed size and conclude how each dropout variant effects the confidence interval.

\paragraph{Contributions}
Our contributions can be summarized as follows:

\begin{itemize}
  \setlength\itemsep{0em}
\item An evaluation of dropout variants and the first application of concrete dropout and curriculum dropout to neural language models. 
\item A novel extension of variational dropout~\cite{gal2016theoretically} to concrete dropout~\cite{gal2017concrete} and curriculum dropout~\cite{morerio2017curriculum} with various schedules.
\item An analysis of test perplexities at each time step, providing an insight as to why and where most errors occur in generating sequences at test time on defacto langauge modelling datasets.
\item An argument for the characteristics of loss surfaces for discrete kronecker delta outputs of high dimensionality where the output space is loosely structured (e.g such as modelling language) and its influence on modelling choices (e.g choosing a curriculum strategy when using curriculum dropout).
\end{itemize}

Before discussing the methodology we give a brief overview of the aforementioned regularization techniques and introduce related research.

\section{Preliminaries}
Dropout is a standard and default regularization technique for many practitioners using neural networks. This involves generating a mask over the weights that is stochastically generated and removes a fraction of neurons ( Bernoulli) or adds noise to the weights ( Gaussian) during learning. This has the effect of preventing neuron units from co-adapting~\cite{hinton2012improving}. Carrying out dropout throughout training can be considered dynamically learning an ensemble of averaged subnetworks which are a child network of the network when $p_d = 0$. Below we briefly describe each variant of dropout we consider using for neural language modelling.

\paragraph{Bayesian Neural Networks and Approximate Inference}
Bayesian Neural Networks (BNNs) are neural networks that place prior distributions $p(\theta)$ over the weights $\theta$ in a neural network $f_{\theta}$ (e.g $\theta = [W_{1}, U_{1}, b_1,...,W_{L}, U_{L},b_L]$ for an LSTM network). The posterior $p(y|\mathcal{D}, \theta)$ can then be computed where $\mathcal{D} \in \{\mathcal{X}, \mathcal{Y}\}$. However, the denominator in Bayes rule \big($p(\theta|\mathcal{D})=p(\mathcal{D}|\theta)p(\theta)\int p(\mathcal{D}|\theta)p(\theta)d\theta\big)$ is in an intractable integral. Therefore the aim is to replace $p(x)$ with a tractable approximate density $q(x)$ from the conjugate exponential family, where $p(\theta)$ is that shown in the denominator. One approximate method for posterior inference that scales relatively well is variational inference (VI).

\paragraph{Variational Inference} VI approximates the posterior by minimizing the distance between a simpler proposal distribution and the true target conditional. Kullbeck-Leibler (KL) divergence is a common measure between such distributions, shown in \autoref{eq:svi}. We cannot minimize the KL directly, but we are able to minimize a function that is equivalent apart from a constant, which corresponds to the Evidence Lower Bound (ELBO), which can be expressed as \autoref{eq:elbo}.

\begin{gather}\label{eq:elbo}
\log p(x) = \log \int_{z} p(x, z) 
= \log \int_z p(x, z) \frac{q(z)}{q(z)} \nonumber \\
= \log \Big( \mathbb{E}_q \Big[ \frac{p(x, z)}{q(z)} \Big] \Big) \geq \mathbb{E}_{q}[\log p(x, z)] - \mathbb{E}_q [\log q(z)]
\end{gather}

Since $p(z|x) = p(z, x)p(x)$ we can express the ELBO in terms of the KL divergence as shown in \autoref{eq:svi} where the $1^{st}$ term is the ELBO and the $2^{nd}$ is the constant term which can be ignored since it does not depend on $q$. Therefore, maximizing the ELBO $ \mathbb{E}_{q} \big[ \log p(x|z) \big] - KL(q(z|x)||p(z))$ is equivalent to minimizing $KL(q(z)||p(z|x))$.

\begin{gather}\label{eq:svi}
KL\big(q(z)||p(z|x)\big) = \mathbb{E}_{q} \log \frac{q(z)}{p(z|x)} 
 = \mathbb{E}_{q}\big[ \log p(z,x) -\log q(z|x)\big] + \log p(x)
\end{gather}

\paragraph{Gaussian Dropout}
~\cite{srivastava2014dropout} proposed using multiplicative Gaussian noise instead of Binary dropout (drawn from a Bernoulli), so $\varepsilon \sim \mathcal{N}(1, \alpha)$ and $\alpha = p/(1-p)$. Since this is a multiplication this can also be interpreted as adding multiplicative noise on $\theta$. This is the equivalent of computing the posterior of the models weights~\cite{kingma2015variational} since the noise allows to sample (consider $w, \theta, \epsilon$ and $\varepsilon$ scalars for the purpose of sampling a single weight) the weight $q(w|\theta, \alpha) = \mathcal{N}(w|\theta, \alpha \theta^{2})$ in the network. Therefore, $w$ is a random variable parameterized by $\theta$ as shown in \autoref{eq:gauss_weight}. This is the same as performing stochastic optimization on the log-likelihood.

\begin{gather}\label{eq:gauss_weight}
	w = \theta\varepsilon = \theta(1 + \sqrt{\alpha}\epsilon) \sim \mathcal{N}(w|\theta, \alpha \theta^{2}) \quad s.t, \quad 
    \epsilon \sim \mathcal{N}(0, 1)
\end{gather}

\paragraph{Variational Dropout}
Variational Dropout (VD) allows the dropout probability to be tuned throughout training, similar to previous work that introduced the idea~\cite{ba2013adaptive} which originally used a binary belief network for learning dropout rates with respect to the inputs. VD uses $q(\theta | \xi, \alpha)$ as the approximate posterior for a model with a simple prior over the variational parameters $\xi$ and $\alpha$ ($\phi = (\xi, \alpha)$) which are both tuned using the aforementioned stochastic variational inference (SVI). The prior  $p(\theta)$ is chosen to be a logscale uniform so that VD with fixed $\alpha$ is the same as Gaussian Dropout (the only prior for which this is the case). Fixing $\alpha$ means $D_{KL}(q(\theta|\xi, \alpha)|p(\theta)$ the ELBO does not depend on $\xi$ and optimization results in maximizing the log-likelihood and using Monte Carlo samples and the KL divergence term $D_{KL}(q(\theta)||p(\theta))$ which is computed analytically or approximated with Monte-Carlo samples. In the context of applying VD to RNNs,~\cite{gal2016theoretically} show that VI can be performed by simply fixing the dropout mask across time-steps. Before this, it was thought that generating masks for each time-step led to a reduction in the signal-to-noise ratio which made backpropogation through time (BPTT) too difficult, particularly in the early stages of training when gradients have higher variance. Hence, it was common that different dropout masks over timesteps were only applied to input and output weights~\cite{zaremba2014recurrent}. \autoref{eq:ll_elbo} shows the log-likelihood of the ELBO for an RNN where $\theta_L$ are the decoder weights at the last layer and $\theta_{1:L-1}$ are the input and hidden layer weights. Since each token of each $N$ sequences are passed through the same function $f_theta$, a sample of $\theta$ is for a whole sequence.

\begin{equation}\label{eq:ll_elbo}
    \mathcal{L}_{mc} \approx \sum_{i=1}^N \log p\Big(y| f_{\theta_{L}}(f_{\theta_{1:L-1}}(x_T, f_{\theta_{1:L-1}}(...f_{\theta_{1:L-1}}(x_1, h_0)...))\big)\Big) + D_{KL}(q(\theta)||p(\theta))
\end{equation}

For the $i^{th}$ row of each weight matrix $w \in \theta$, variational parameter $m$ (updated with gradient descent) and given dropout probability $p_d$ the approximate distribution is $q(\theta) = p_d\mathcal{N}(w_i; 0, \sigma^{2} I) - (1-p_d)\mathcal{N}(w_i; m_i, \sigma^{2} I)$. When using RNN gated networks, $m$ can be learned for multiple $w \in \theta$ to speed up computation. However, we use untied weights to avoid the performance degradation in this work.

~\cite{molchanov2017variational} use VD to tune individual dropout rates per neuron, showing good generalization performance by promoting more sparsity in neural networks, namely Sparse Variational Dropout (SVD). Lastly, ~\cite{hron2017variational} have argued that the log-uniform prior used in Variational (Gaussian) Dropout is not understood in terms of the Bayesian context as it is not produce a proper posterior. Therefore, the reported sparsity benefits are due to a group of non-Bayesian approaches. We note this as we consider the GVD and variational extensions to curriculum dropout.

\paragraph{Concrete Dropout}
Concrete dropout~\cite{gal2017concrete} also allows $p_d$ to be learned using the concrete distribution which is a continuous relaxation $\tilde{z}$ of the discrete random variables (e.g Bernoulli dropout) by re-parameterizing the distribution. This is achieved by using a pathwise derivative estimator which takes $\theta$ as $g(\theta, \epsilon)$ where $\epsilon$ is a random variable independent of $\theta$. Instead of drawing uniformly as in the Bernoulli case, a softmax is used to draw from the continuous distribution with a temperature $\tau$ controlling the kurtosis of the softmax and Stochastic Gradient Descent (SGD) can be used for optimization.

\autoref{eq:concrete_dropout_1} shows the objective function where the continuous relaxation $\tilde{z}$ on the Bernoulli mask $z$ (equivalent to aforementioned $w$ which is the Gaussian equivalent), $\sigma$ is the sigmoid function and, $u$ is drawn uniformly between 0-1, $\tau$ is the temperature and  $\epsilon$ is a lower bound on $p_d$ to prevent $- \infty$ outputs in the log terms ($\tau=0.1$ and $\epsilon=1e-6$ in our experiments). The pathwise derivative estimator is then used to obtain the continuous estimate $\tilde{z}$ of $z$.

\begin{equation}\label{eq:concrete_dropout_1}
	\tilde{z} = \sigma\Bigg(\frac{\log\Big(\frac{p_d + \epsilon}{1-p_d + \epsilon}\Big) + \log\Big(\frac{u + \epsilon}{1-u + \epsilon}\Big)}{\tau}\Bigg)
\end{equation}

\paragraph{Curriculum Dropout}
~\cite{morerio2017curriculum} propose to use a time scheduled dropout  where noise is added to the input and hidden layer inputs increases over time, incrementally increasing the optimization difficulty given the increase in stochasticity. Concretely, $\theta = (1 - \bar{\theta})\exp(\gamma t) + \bar{\theta}, \; \gamma > 0$ where $\bar{\theta}$ where $\bar{\theta}$ is an upper bound in $0.5 \leq \theta \leq 0.9$ as described in the original dropout paper~\cite{srivastava2014dropout} where layers nearer the output have smaller $p_d$. This is motivated by the fact that neuron co-adaptation mostly occurs later in training, whereas label fitting is happening throughout the earlier epochs where co-adaptation is less of an issue. Moreover, we agree with the intuition expressed in the paper that neuron co-adaptations become a problem for overfitting later in training, while in the early stages of training, co-adaptations can reveal some structure by the parameters self-organizing that tend towards an optimal configuration. Hence, this can be reflected in the change of $p_d$ in curriculum dropout.

\section{Related Work}

Recently, there has been few papers that focus on evaluations of NLMs in terms of hyperparameter optimization and network comparisons for identifying high performing networks and optimal NLM parameter configurations. ~\cite{melis2017state} have recently reevaluated various neural language models both in architecture and regularization methods using black-box hyperparameter tuning provided by Google Vizier~\cite{golovin2017google} which uses batched GP bandits via the expected improvement acquisition function~\cite{desautels2014parallelizing}. They conclude that plain LSTMs achieve SoTA compared to more complex architectures when properly regularized. 

Similarly, ~\cite{merity2017regularizing} have also investigated the effects of different regularization and optimization on LSTM language models. Weight tying, weight dropping, randomized backpropogation through time (BPTT), activation regularization (AR) and temporal AR (TAR) were also tested in their experiments, while applying a continuous cache pointer to NLMs which improved performance (originally proposed by ~\cite{grave2016improving}). 

In contrast to our work, the focus of the aforementioned papers was evaluation on the dropout configurations (input, intra-layer and/or output dropout) but does not include dropout techniques that consider adapting/learning the probability during training (e.g concrete dropout or curriculum dropout). Moreover, we include an analysis of the confidence of each prediction at test time to gain a better understanding of where the models tend to perform poorly for these alternative dropout techniques.

\section{Methodology}
\subsection{Loss Space Curvature}\label{sec:lsc}

\begin{wrapfigure}{R}{7cm}
\vspace{-3em}
    \includegraphics[scale=0.5]{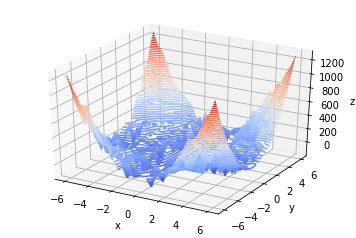}
    \caption{A synthetic illustration of perplexity (z-axis) loss surface for language modeling (e.g PTB) for 2 parameters (x-y axis)}\label{fig:ppl_loss_surface}
    \vspace{-1em}
\end{wrapfigure}

Before introducing the dropout variants we set the premise that when modelling a high dimensional dirac distribution where the input frequency distribution of $X$ follows a power law (i.e transitions are far more ubiquitous for common words and very sparse for rare words), the error surface approximately forms a truncated cone shape where saddle points and local minima are ubiquitous towards the bottom (see illustration in \autoref{fig:ppl_loss_surface}). Large improvements are made on validation perplexity in the early to intermediate stages where the model loss decreases steeply along the slope of the \say{inverted wine bottle}-like curvature ~\cite{anandkumar2016efficient}. 
\autoref{fig:ppl_loss_surface} illustrates an example this type of loss surface\footnote{For viewing purposes we only show high perplexity at the extremums of $x$ and $y$ so the loss surface at the bottom can be viewed.}. Although, we acknowledge that every hyperparameter and optimization choice effects the shape of the loss landscape.

\subsection{Model Description}
For testing the dropout variants we use the log-likelihood setting for training LSTM, GRU and Highway network language models. For a sequence pair $(X_{1:T}, Y_{1:T})$ of length $T$, at time $t$, $x_t$ is passed to a parametric model along with the previous hidden state vector $h_{t-1}$ to produce an output from its last layer $\tilde{h} =f_{\theta}(x_t, h_{t-1})$. Then, the probability distribution $\hat{y} = \phi(\tilde{h}\cdot W)$ is computed where $\phi$  is a softmax function, $W$ are the decoder weights and the predicted token is retrieved via the $\arg\max$ operator. During training the batches are grouped and aggregated by length for more efficient mini-batch processing. To signify the start and end of a sequence for an input $X$, we use $\mathtt{<sos>}$ and $\mathtt{<eos>}$ tags respectively. Standard log-likelihood training is used as shown in \autoref{eq:log_prob}.    

\begin{equation}\label{eq:log_prob}
\frac{1}{T}\sum_{t = 1}^{T} \log p(y_t|y_{1:t-1}, X; \theta)
\end{equation}

This involves using a cross-entropy loss and SGD for optimization. In this paper, we use an annealed learning rate based on the validation perplexity scores. This means that if the validation perplexity is worse at the epoch $i$ than it was at epoch $i-1$, we reduce $\alpha_{i+1} = \alpha_{0}\exp(-c)$ where $c$ is a counter that increments by 1 every time the validation performance decreases throughout training. At test time, we generate a sequence by sampling from the models output via a multinomial distribution.

\begin{equation}
	\theta_i: = \theta_i - \alpha_i \tilde{\nabla} \quad s.t \quad \alpha_i \leq \alpha_{i-1} \; \forall i 
\end{equation}

\subsection{Dropout Details}
We make two relatively simple extensions of dropout for NLM. First, we adapt curriculum dropout to RNNs by fixing the mask over timesteps, extending VD~\cite{kingma2015variational} to incorporate curriculum schedules. The dropout rate $p_d$ is then monotonically increased throughout training, inversely proportional to the validation perplexity. This is motivated by the observation that large improvements are made early in training on language modelling datasets and more generally we suspect this to be the case for problems with similar input distributions, as mentioned in \autoref{sec:lsc}.  We test various schedules for the increase including a linear, sigmoidal and an exponential increase such as that shown in \autoref{eq:update}, where $N$ is the number of epochs. For the experimental results we use fixed-dropout masks (i.e the same as VD) for Curriculum Dropout with all schedules up to $p_d=0.3$ beginning at $p_d=0$. Likewise, we fix the mask with concrete dropout for all configuration (input, hidden and output and all). 

\begin{equation}\label{eq:update}
p_d:= p_d + |p_d - (2^{i/N}-1)| \quad \forall i \in N
\end{equation}

We also carry this out for concrete dropout. For the experimental results we use fixed-dropout masks (i.e the same as VD) for Curriculum Dropout with exponential, sigmoid and linear schedules up to $p_d=0.3$ beginning at $p_d=0$. Likewise, we fix the mask with concrete dropout for all configuration (input, hidden and output and all). We choose regularization of the parameters in concrete dropout to have a weight of $0.1$ in the loss. We found during training that lower levels of regularization drove $p_d$ near to 0 since early in training the language model has a high learning rate which results high variance in the gradient, that in turn results little influence of the learned dropout probability since the regularization is given too small of a weight. In contrast to curriculum dropout where we begin with $p_d=0$ that is monotonically increasing to $0.3$, concrete dropout begins at $p_d=1$ and decreases throughout training before stabilizing to an approximately optimal $p_d$.

\subsection{Training Details}\label{sec:training_details}
For all experiments we use (relatively small) 2-hidden layers for all networks with an embedding size of $e_s = 300$ for both PTB and WikiText-2 respectively. For optimization we use annealed SGD where the learning rate is halved if validation perplexity is worse than its previous settings. The embedding input for each of the models are initialized uniformly at random $[-0.1, 0.1]$. For PTB and WikiText-2, training is run for 40 epochs which sufficed for convergence. We choose a mini-batch size of $64$ with truncated backpropogation set to $30$ timesteps. Gradient norms are clipped above $0.3$ (in an effort the mention saddle points) and the hidden state from the previous batch is used to initialize the current batch. We use annealed SGD where the learning rate begins at $10$ and is multiplied by $0.3$ if there validation perplexity increases compared to the previously computed validation perplexity.

\section{Results}\label{sec:results}
\subsection{Penn-Treebank Analysis}
\autoref{tab:ptb_dropout_results} shows dropout results for a dropout rate $p_d$ for all models, on all dropout variants for Penn-Treebank. For static dropouts $p_d=0.2$ during training, $p_d$ is learned for concrete dropout and $0 \leq p_d \leq 0.3$ monotonically increasing for curriculum dropout. The 3 dropout variants CSigmoid, CExp and CLinear correspond to schedules for Curriculum (C) dropout with fixed masks across time-steps, like variational dropout. Shaded cells correspond to the best performance for each block of results.

For PTB language modelling, we find best LSTM results are obtained using a time-fixed curriculum dropout using an exponential increase (CExp). For GRU and Recurrent Highway Networks, we find a linear increase (CLinear) slightly outperforms other schedules. We find that Concrete Dropout performs relatively well, particularly for the GRU Network which is competitive with Variational Curriculum Dropouts.

\begin{table}[ht]
\centering
\captionsetup{justification=centering, margin=0cm}
\resizebox{.65\linewidth}{!}{%
\begin{tabular}{c|cccc}
\toprule
($p_d =0.2/0.3$) & Input & Output & Hidden & All\\
LSTM & \multicolumn{4}{l}{} \\
\midrule
Standard & 94.68/86.27 & 86.19/79.24 & 86.37/78.96 & 83.48/76.53\\
Gaussian & 93.45/85.81 & 85.71/80.20 & 85.45/78.12 & 82.02/75.89\\
Variational & 93.13/85.97 & 85.45/79.88 & 85.30/79.94 & 81.83/75.71\\
Concrete & 93.22/85.91 & 85.08/79.17 & 85.19/78.20 & 81.35/75.64\\
CSigmoid & 86.56/85.31 & 85.08/84.69 & 85.44/84.82 & 83.65/77.01\\
CExp. & \cellcolor{black!20}86.11/80.01 & \cellcolor{black!20}82.78/79.85 & \cellcolor{black!20}81.20/78.28 & \cellcolor{black!20}
81.08/75.28\\
CLinear & 88.11/87.31 & 88.13/87.47 & 86.20/83.94 & 87.12/83.80\\
\midrule
GRU & \multicolumn{4}{l}{} \\
\midrule
Standard & 95.68/89.76 & 91.45/86.12 & 92.49/86.58 & 87.73/81.76\\
Gaussian & 95.84/88.86 & 89.78/85.90 & 92.25/86.74 & 87.58/81.43\\
Variational & 95.31/89.09 & 93.21/86.15 & 90.04/86.24 & 86.50/80.40\\
Concrete & 87.11/83.68 &86.04/80.35 &87.38/81.70 & 87.28/80.31\\
CSigmoid & 94.00/88.42 & 88.61/85.88 & 95.23/90.02 & 85.51/80.70\\
CLinear & \cellcolor{black!20}92.68/90.38 & \cellcolor{black!20}87.05/83.31 & \cellcolor{black!20}92.31/90.60 & \cellcolor{black!20}85.52/79.81\\
CExp. & 92.40/88.75 & 88.39/93.85 & 91.37/87.26 & 86.85/81.55\\
\midrule

Highway & \multicolumn{4}{l}{} \\
\midrule
Standard & 102.91/88.56 & 93.21/79.09 & 94.56/85.51 & 92.02/83.93\\
Gaussian & 103.65/88.52 & 94.29/79.86 & 94.29/79.77 & 93.12/82.40\\
Variational & 100.91/87.11 & 92.41/78.11 & 93.58/82.24 & 91.20/81.44\\
Concrete & 101.63/86.11 & 92.29/79.86 & 93.31/82.12 & 91.33/81.16\\
CSigmoid & 99.68/81.86 & 91.71/78.02 & 92.60/82.49 & 89.56/80.27\\
CLinear. & \cellcolor{black!20}97.12/80.25 & \cellcolor{black!20}89.31/77.38 & \cellcolor{black!20}91.34/82.02 & \cellcolor{black!20}88.72/79.31\\
CExp & 101.48/82.69 & 93.01/79.97 & 93.27/83.27 & 90.52/81.10\\
\bottomrule
\end{tabular}%
}
   \caption{(Validation/Test) Results of Dropout Variants for PTB NLMs}
  \label{tab:ptb_dropout_results}
\end{table}

\begin{wrapfigure}{R}{8cm}
    \centering
    \captionsetup{justification=centering}
    \includegraphics[scale=0.5]{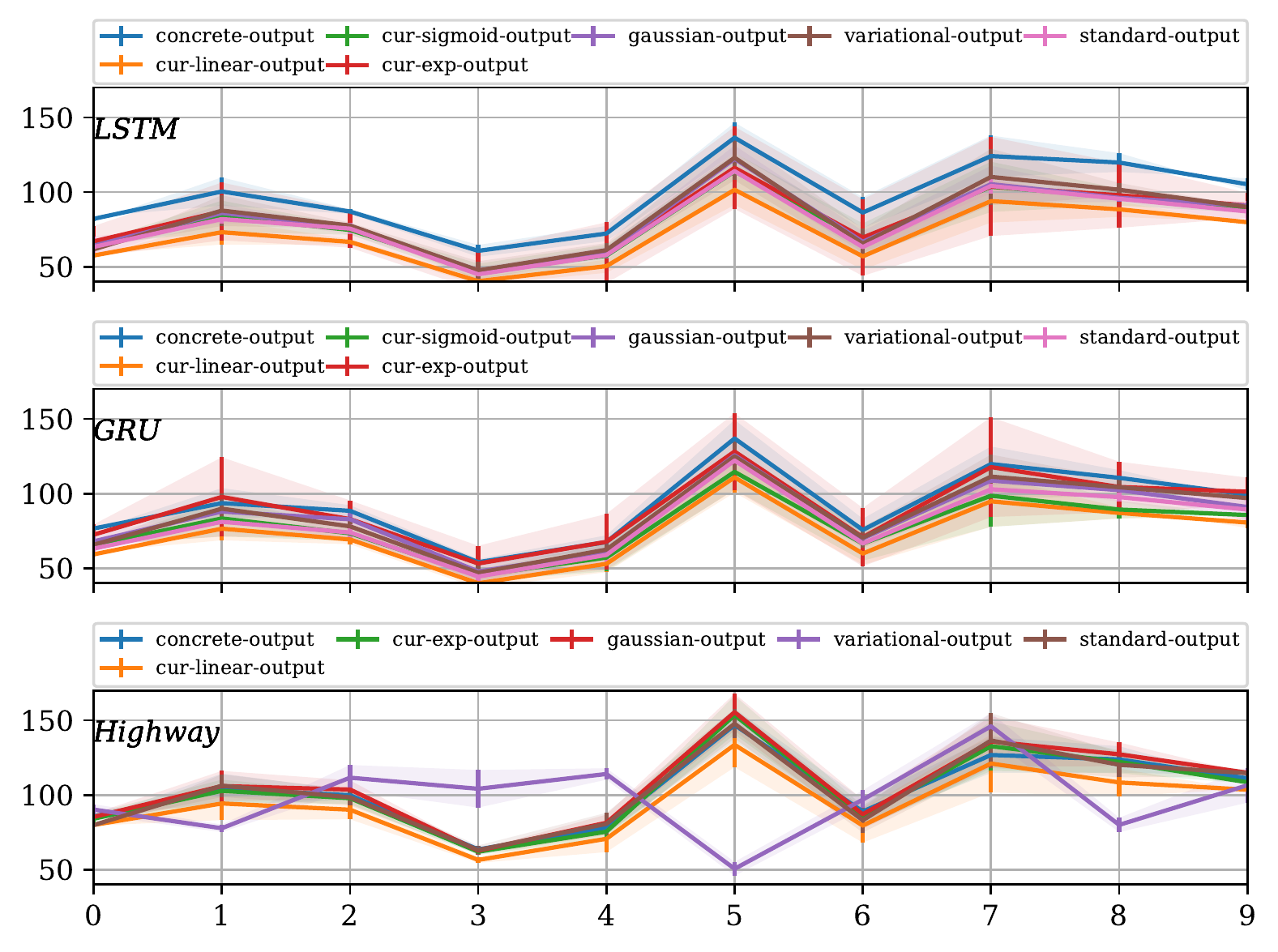}
    \caption{Penn-Treebank dataset - mean test perplexity per time-step. (The confidence bars are standard deviation in perplexity)}\label{fig:ptb_uncertainty_plot}
    \vspace{-1em}
\end{wrapfigure}

\autoref{fig:ptb_uncertainty_plot} shows the log-likelihood loss 
varies across time-steps for a sequence length $T=10$ when using the dropout techniques on the decoders outputs\footnote{For brevity, we focus on decoder dropout as it has results in the most change in performance of all variants}. The figure shows the mean and average standard deviation in loss at each time step $t$ in the test set. We see early on that the loss variance is slightly higher in standard dropout than it is compared to concrete dropout. This suggests that learning the dropout rate during training does reduce the loss variance when the model is required to generate a sequence at test time, hence better mitigating the variance in accumulated errors along the generated sequence.

\paragraph{PTB - Test Perplexity Per Time-Step}
To understand and pinpoint where the models perform well or poor at test time, we analyse the perplexity from the mean cross-entropy loss for each time step along with the lower and upper mean absolute deviations (MAD). This allows us to identify on average, which timesteps are the most difficult for the model to perform well on for each respective dataset. We assume this to be the confidence interval (CI) instead of the prediction interval (PI) by allowing $\mathbb{E}[Y|X] \approx \mathbb{E}[\mathcal{Y}|\mathcal{X}]$ as we do not know all transitions that come from the respective discrete input-output sets $\{\mathcal{X}^{\mathbb{Z}}, \mathcal{Y}^{\mathbb{Z}}\}$.

One common trend between all models is that when the loss is considerably reduced at $t$ (e.g $t=5$ for all networks), it is countered with an increase cost within the next timesteps. If a good prediction is made, it is soon after penalized. This would be intuitive for rare words, as correctly predicting a relatively rare word would be more difficult to recover from as it has seen less transitions in the training set. This is mildly surprising considering PTB has no rare words due to tokenization. We also note that there does not seem to be strong evidence for exposure bias, which again we postulate that this is because of the preprocessing steps undertaken in PTB. 

\subsection{WikiText-2 Analysis}
For WikiText-2, we find best results overall are obtained using VCD with an exponential schedule for an LSTM and a linear increase for GRU and Highway networks. All VCD schedules show results that are close and outperform static dropout rates. These dropout methods are particularly effective when used on the output. We suspect this is because when compounding errors occurs, using log-likelihood without any dropout in the networks makes it difficult for the model to re-correct early mistakes made in the sequence. By using dropout, the added stochasticity throughout training allows the model to seek alternative transitions where it had previously made mistakes. Since most of the difficult transitions are that which include rare words, we would expect that near convergence most of the loss reductions are found in these predictions. Hence, smoothing the loss surface near the end of training is critical, which is why we opt for using VCD.

\begin{table}
\centering
\captionsetup{justification=centering, margin=0cm}
\resizebox{.65\linewidth}{!}{%
\begin{tabular}{c|cccc}
\toprule

($p_d$ =0.2/0.3) & Input & Output & Hidden & All\\
\textit{LSTM} & \multicolumn{4}{l}{} \\
\midrule
Standard & 154.01/139.30 & 153.33/138.97 & 150.02/137.11 & 149.77/136.82\\
Gaussian & 154.21/140.49 & 156.01/143.16 & 155.39/143.91 & 155.53/142.08\\
Variational & 153.23/138.80 & 152.22/138.68 & 152.40/138.95 & 149.31/136.74\\
Concrete & 154.25/139.50 & 153.29/138.03 & 150.41/136.26 & 148.13/134.57\\
CSigmoid & 153.10/139.81 & 141.39/126.66 & 143.28/134.30 & 139.27/127.08\\
CLinear. & 153.04/139.99 & 145.11/137.93 & 148.10/137.81 & 144.68/126.31\\
CExp & \cellcolor{black!20}152.14/138.60 & \cellcolor{black!20}151.73/137.25 & \cellcolor{black!20}139.24/135.31 & \cellcolor{black!20}137.36/125.54\\
\midrule

\textit{GRU} & \multicolumn{4}{l}{} \\
\midrule
Standard & 156.57/142.12 & 154.72/141.43 & 153.97/140.89 & 154.23/140.40\\
Gaussian & 154.99/140.33 & 154.29/139.26 & 152.49/139.29 & 153.55/139.46\\
Variational & 150.42/138.38 & 149.03/137.44 & 150.32/137.84 & 149.30/138.98\\
Concrete & 157.52/150.17 & 160.38/151.64 & 157.24/148.80 & 159.02/149.96\\
CSigmoid & 154.08/140.22 & 137.52/123.59 & 139.59/124.37 & 139.45/127.73\\
CLinear & \cellcolor{black!20}151.32/138.05 & \cellcolor{black!20}138.32/124.72 & \cellcolor{black!20}128.16/125.21 & \cellcolor{black!20}130.02/125.41\\
CExp. & 156.27/142.12 & 128.10/126.45 & 149.88/121.71 & 146.33/128.41\\
\midrule

\textit{Highway} & \multicolumn{4}{l}{} \\
\midrule
Standard & 160.98/150.25 & 162.74/149.22 & 158.71/146.51 & 158.49/146.40\\
Gaussian & 159.31/148.79 & 159.05/148.69 & 158.29/145.74 & 158.02/145.64\\
Variational & 155.25/146.69 & 159.33/142.39 & 151.34/143.09 & 152.96/142.55\\
Concrete & 158.34/149.46 & 159.18/143.76 & 153.21/144.01 & 153.39/143.14\\
CSigmoid & 154.33/144.71 & 152.74/139.48 & 150.06/142.46 & 151.13/140.93\\
CLinear. & \cellcolor{black!20}153.98/144.87 & \cellcolor{black!20}151.38/138.70 & \cellcolor{black!20}150.25/139.89 & \cellcolor{black!20}149.25/139.23\\
CExp & 154.02/145.14 & 152.11/139.46 & 151.08/142.71 & 151.42/140.57\\
\bottomrule
\end{tabular}%
}
   \caption{WikiText-2 NLM Dropout Variant (Validation/Test) Results}
  \label{tab:wiki2_dropout_results}
\end{table}

\begin{wrapfigure}{R}{8cm}
    \centering
    \captionsetup{justification=centering}
    \includegraphics[scale=0.5]{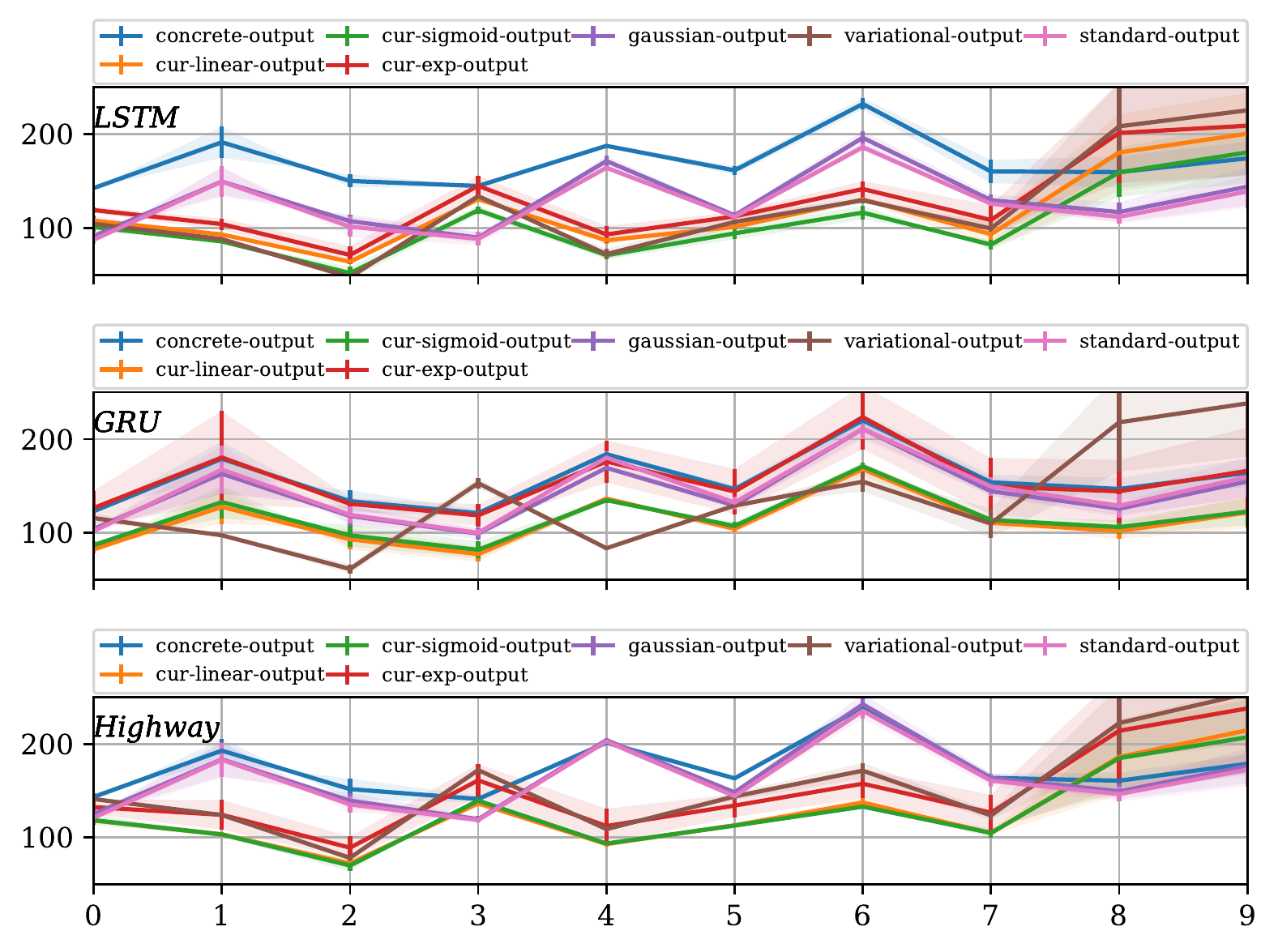}
    \caption{Mean Test Perplexity on WikiText-2}\label{fig:wiki2_uncertainty_plot}
        \vspace{-1em}

\end{wrapfigure}

\paragraph{WikiText-2 - Test Perplexity Per Time-Step}
In contrast to PTB, \autoref{fig:wiki2_uncertainty_plot}, we do in fact see more evidence for compounding errors, particularly towards the end of the test sequence at $t=[8-10)$.  This has also been noted and addressed in prior work~\cite{neill2018curriculum} and coincides with the intuitive idea that it is more likely when terms which fall within the long-tail of the unigram distribution are included and not discarded via preprocessing steps (e.g stemming). This also means that towards convergence the loss surface contains more local minima and saddle points than if we only considered tokens which were more frequent. Again, we find that curriculum dropout with a linear schedule has shown some improvement over using static dropout. Early in training there is no overfitting and hence there is no necessity for regularization, but nearer convergence the increase in dropout rate has the effect of smoothening the loss surface, as discussed in \autoref{sec:lsc}. 

\paragraph{Observations}
We found that when using concrete dropout regularization had to be set relatively high. This is because the decrease in perplexity is quite steep in the early stages in training which corresponds to a decrease in the learned dropout rate. This results in $p_d$ being close to 0 before the NLM is near convergence, which where the chosen dropout rate is most critical. We also considered using a schedule for the amount of regularization which was proportional to the change in validation perplexity. By setting regularization high, we found the dropout rate converged to $p_d=0.26$ which is near to default of $p_d=0.3$ used for the static dropout rates.

Large improvements are made early since most of the perplexity decrease during training is due to the ubiquity of stopwords and terms that are highly frequent in the power law distribution, which are relatively easier to classify correctly. 

The difficult part is correctly predicting rare tokens, or more precisely, tokens that fall into the long tail of the power law distribution, as mentioned in \autoref{sec:lsc}. Subsequently, poor performance on these tokens can lead to worse performance on common tokens due the problem of exposure bias. 

We also find that a trend in compounding errors is more evident for WikiText-2 which does not discard words with low frequency. Hence, making an unseen transition is more likely, in constrast, making a poor prediction on PTB early on is recoverable since even a bad prediction can lead to future correct states given the likelihood of observing such transition in the dataset is more likely. This is reflected in the fact that the test perplexity has a higher variance per time-step on WikiText-2 compared to PTB. Interestingly, we also see that using variational dropout improves performance but when compounding errors occur it is more difficult to recover from. In contrast, when the random mask changes over time steps, this added stochasticity prevents the fast increase in perplexity, as seen across each model from \autoref{fig:wiki2_uncertainty_plot}. 

\section{Conclusion}
This paper has compared dropout variants, which (1) keep the dropout rate static during training and (2) those which learn the dropout rate or adapt the rate using a chosen schedule. We find that adapting curriculum dropout can improve performance, particularly when modelling raw text which does not carry out any preprocessing steps such as word removal or canonicalization.
We have introduced a variational curriculum dropout variant whereby the random mask is fixed across timesteps which has shown to outperform the its static dropout rate counterparts. In general, we find each tested schedule to be comparable but a linear schedule obtains the best results overall. We also extend concrete dropout to RNNs in the same way (i.e fixing the mask across time-steps). Lastly, we find that compounding errors are, intuitively, most evident when words in the long-tail of the unigram distribution are not removed. This is reflected in perplexity deviation over time-steps and the corresponding deviations from the mean in the test set.

In summary, we encourage the use of adaptive dropout rates with fixed dropout masks across time-steps and to carefully consider the performance impact of preprocessing datasets for many natural language tasks, including language modelling. 

\bibliography{iclr2019_conference}
\bibliographystyle{iclr2019_conference}

\end{document}